\documentclass{article}
\usepackage{amssymb}
\usepackage{tikz}
\usepackage{url}
\usepackage{authblk}

\title{XSTEM: An exemplar-based stemming algorithm\thanks{I was introduced to exemplar theory by Keith Johnson in a phonetics seminar at Ohio State. His model is called XMOD, which, as he notes, is the best name \cite{Johnson2005}. The name XSTEM comes from that.}}
\author{Kirk Baker}
\affil{Lexical Intelligence\vspace{-0.5em}}
\affil{The Analytics Research Institute}

\date{May 10, 2022}
\begin{document}

\maketitle

\begin{abstract}
Stemming is the process of reducing related words to a standard form by removing affixes from them. Existing algorithms vary with respect to their complexity, configurability, handling of unknown words, and ability to avoid under- and over-stemming. This paper presents a fast, simple, configurable, high-precision, high-recall stemming algorithm that combines the simplicity and performance of word-based lookup tables with the strong generalizability of rule-based methods to avert problems with out-of-vocabulary words.
\end{abstract}

\section{Introduction}
Stemming is the process of reducing related words to a standard form by removing affixes from them. For example, \textit{eating}, \textit{eats}, and \textit{eaten} can be reduced to the standard form \textit{eat} by removing the \texttt{-ing}, \texttt{-s}, and \texttt{-en} from each. Stemming is a foundational step in many text processing pipelines, including information retrieval and language modeling \cite{BRYCHCIN201568}, and as a feature reduction step for classification or lexical transfer learning tasks \cite{Agrawal2022}.

Most stemming algorithms are either rule-based or corpus-based. Rule-based stemmers use a set of manually created rules to transform words to their base form, typically in conjunction with reference to a dictionary for handling exceptions or modulating their output in some fashion. Corpus-based stemmers typically employ statistical machine learning methods to equate word forms based on distributional regularities derived from large amounts of text. Although researchers have demonstrated impressive results with corpus-based approaches (e.g., \cite[and references therein]{BRYCHCIN201568}), rule-based stemming implementations are more widely available through software distributions such as Apache Lucene\footnote{\url{https://lucene.apache.org}} or Python NLTK\footnote{\url{https://www.nltk.org}}.

Stemming algorithms differ with respect to whether they aim to produce a real-word form (often called a \textit{lemma}) or simply a truncated version of the input (often called a \textit{root}). For example, a word-oriented stemmer might produce \textit{polite} as the output for \textit{politely} whereas a truncating algorithm might produce \textit{polit}. Common problems encountered by stemming algorithms are over-stemming and under-stemming. Over-stemming occurs when unrelated words are reduced to the same form, such as reducing both \textit{rations} and \textit{rational} to \textit{rat}. Under-stemming occurs when related words are reduced to distinct forms, such as reducing \textit{recognize} to \textit{recogn} and \textit{recognition} to \textit{recognit}. Another common problem is failing to handle words that are not contained in a referenced lexical resource. 

This paper presents a rule-based, word-oriented stemming algorithm for English that is simultaneously high precision and high recall. It does not over- or under-stem, handles a wide range of irregular word classes, and generalizes accurately to unknown words. The algorithm is fast, straightforward to implement, easy to configure for different levels of stemming, and comes with an open-source reference implementation. The stemmer described herein overcomes each of the problems noted for previous approaches in the section below.

\section{Related Work}
This section describes previous approaches to English stemming that are widely used today or were fundamental to the development of those widely used algorithms. 

\paragraph{Lovins Stemmer}
The Lovins stemmer uses multiple steps to remove suffixes, rewrite truncated forms, and conflate similar forms. Each stage relies on a set of ordered rules (294 suffix rules, 29 conditions by which they are applied, 34 transformation rules, and 17 conflation rules) to produce a final result \cite{Lovins1968,Hull96,Jivani2011,Moral2014ASO}. Suffixation rules are based on the longest match found. The Lovins algorithm is an aggressive stemmer (e.g., \textit{nationally} stems to \textit{nat} \cite{Hull96})  and does not aim to produce real-word forms. It has been noted to improperly conflate distinct words (e.g., \textit{probe} and \textit{probate} are respectively stemmed to \textit{probe} and \textit{prob}, and then conflated by partial-matching \cite{Moral2014ASO}).

\paragraph{Dawson Stemmer} The Dawson stemmer \cite{Dawson74} extends the work of Lovins by including a larger set of about 1200 suffixes \cite{Jivani2011}. Suffixes are reversed-indexed by length and last character. Like Lovins \cite{Lovins1968}, the Dawson stemmer is a single-pass algorithm that removes the longest matching suffix in a single step and recodes the remaining stem into a valid word using a mapping table  \cite{Oreilly}. Previous authors have noted that the Dawson algorithm is complex to implement and lacks a reference implementation  \cite{Hull96, Jivani2011}.

\paragraph{Porter Stemmer} The Porter algorithm successively removes suffixes from a predefined set using a multi-step approach \cite{Porter80,Krovetz93,Jivani2011}. Within each of the algorithm's five steps, suffix rules are tried until one is accepted and the next step is applied to the output of the previous step. The resulting truncated form is returned at the end of the fifth step. The Porter algorithm also does not aim to produce real-word forms, and may conflate words with different meanings to the same stem while shortening related words to distinct stems. For example, \textit{generous}, \textit{generalization}, and \textit{generic} all stem to \textit{gener}, while \textit{recognize} and \textit{recognition} are stemmed to \textit{recogn} and \textit{recognit}, respectively \cite{Hull96}.

\paragraph{Krovetz Stemmer} The Krovetz stemmer \cite{Krovetz93} modifies the Porter algorithm to check words against a dictionary prior to performing each suffixation step. If the output of a prior step is found in the dictionary, the algorithm terminates and the previous output is returned. For example, if the dictionary contains \textit{generalization}, the stemming of the plural form \textit{generalizations} will terminate after removal of the final \texttt{-s} rather than proceeding to subsequent steps that would result in further truncation. For word forms that are not in the dictionary, the Krovetz stemmer generally returns heavily stemmed output resembling the output of the Lovins algorithm. The Krovetz stemmer is explicitly designed to return complete words and to be highly configurable, providing six supplemental files (exceptions, countries, proper nouns, primary dictionary, supplemental dictionary, and direct conflations) that can be customized for desired stemming results \cite{Kstem}. 

The main drawback of the Krovetz stemmer is its inability to handle words that are not in the lexicon \cite{Hull96}. Because its lookup tables rely on full-word matching, it is unable to generalize to classes of desired stemming behavior, instead requiring each specific example to be recorded. For example, the Krovetz algorithm reduces \textit{viruses} to \textit{viruse}, which means that even after stemming, the singular and plural forms of \textit{virus} would fail to match. Although we can configure a mapping from \textit{viruses} to \textit{virus}, this configuration will not modify the output for compound words ending in \texttt{-virus}, such as \textit{retroviruses}, \textit{adenoviruses}, \textit{coronaviruses}, etc. A PubMed title search\footnote{\url{https://pubmed.ncbi.nlm.nih.gov}, accessed 30 April 2022.} for \textit{*viruses} yields 450 distinct compounds of this type, illustrating the limitations of dictionary-based stemming methods.

\paragraph{Paice Stemmer}
The Paice stemmer \cite{Paice90} is an iterative, rule-based stemmer that groups rules by the final character of the suffix to which they apply. For example, rules pertaining to words ending in \texttt{-s} form a group; rules applying to words ending in \texttt{-ing} are grouped under the letter \texttt{g}, and so forth. Each rule consists of the following components:
\begin{itemize}
	\item the suffix to be matched, stored in reverse order
	\item the number of characters to remove, which may be zero
	\item the new suffix to be appended, which may be null
	\item a continuation symbol, which indicates whether to continue stemming the intermediate form
	\item an optional flag that controls stemming of intact vs. partially stemmed word forms
\end{itemize}
In addition to these rules, the Paice stemmer requires an additional set of constraints that limit improper stemming of certain words such as \textit{rant}, \textit{rice}, and \textit{river}, among others, although Paice notes that a lookup table may be a preferable solution \cite[p.~58]{Paice90}.

\section{Comparison with Previous Work}
This section describes our proposed stemming algorithm and compares it to previous work, showing how it builds upon those earlier methods and mitigates problems noted for each. 

XSTEM is a fast, configurable, high-precision, high-recall stemming algorithm that combines the performance and interpretability of word-based lookup tables with the strong generalizability of rule-based methods. Like Lovins \cite{Lovins1968} and Dawson \cite{Dawson74}, XSTEM is based on the longest-matching suffix principle. Suffix rules are stored in a reverse character trie, providing a highly performant lookup structure that is equivalent to the constant runtime performance of retrieving hashed strings from a map. Dawson \cite{Dawson74} and Paice \cite{Paice90} utilized similar reverse storage mechanisms to limit the number of iterations within an applicable class of rules. The components of an XSTEM suffix rule are analogous to those of Paice, and specify 1) the suffix to match, 2) the number of characters to remove from the end of the input (this number may be zero), and 3) a replacement suffix to be appended (this replacement may be null).

Like Krovetz \cite{Krovetz93}, XSTEM produces valid words as output. Unlike Krovetz, XSTEM does not rely on a lexicon\footnote{It does have exceptions, similar to Paice and Krovetz; these are detailed in Section 4.} to determine its stemming behavior. By eliminating the dependency on a dictionary file as a requirement for producing real-word output, XSTEM overcomes the out-of-vocabulary problem documented for Krovetz in \cite{Hull96}. Rather than relying on strict lexical lookup, XSTEM generalizes from a pool of exemplar suffixes to handle regular and irregular forms with equal proficiency.

Also like Krovetz, but in contrast to Dawson \cite{Hull96, Jivani2011}, XSTEM is modular and highly configurable. For the Krovetz stemmer, these properties are most notably reified by its supplemental vocabulary files, which are separated into lexical classes (such as countries, proper nouns, and direct conflations) for ease of modification \cite{Krovetz93}. XSTEM follows this paradigm and utilizes a list of proper nouns (names whose structure otherwise makes them eligible for stemming, such as \textit{Denning} or \textit{Maldives}) and a list of exceptions (primarily short words that would otherwise be ineligible for stemming, such as \textit{tied/tie} or irregular verbs, such as \textit{brought/bring}). The exceptions list is relatively short; the majority of irregular suffixes are handled by the suffix trie and do not require separate configuration. Finally, XSTEM requires a specification of suffix rules, which are also contained in an editable text file. This file currently contains approximately 1500 rules; each rule consists of either one, two, or three fields representing the suffix to match, the number of characters to remove (if any), and the new suffix to be appended (if any).

Related to its configurability, XSTEM is implemented as a multi-pass stemming algorithm, a feature shared with Porter \cite{Porter80} and Paice \cite{Paice90}. Each step outputs real-word forms, and any given step is optional. This design allows XSTEM to be configured as a light stemmer (i.e., plural only) to a more aggressive one with minimal effort. Compared to previous iterative rule-based approaches, XSTEM's algorithm is simple: a lookup function performed by trie traversal. Rules which attempt to capture the subtlety and complexity of English stemming behavior, which is a mishmash of productive and semi-productive processes and historic borrowings from other languages with their own idiosyncracies, are complex and arbitrary, typically referring to specific combinations of consonants and vowels, syllable struture, or specific characters in certain positions to make the rule work. In our view, it is only possible to write such rules by gathering and examining sufficient data and counterexamples to refine them, but it is more productive to omit the intermediate step of converting exemplars to rules and utilize a model that generalizes from the exemplars directly.

To summarize, the overall advantages of XSTEM compared with previous approaches are:
\begin{itemize}
	\item it is fast, with performance equivalent to direct lookup methods
	\item it is configurable, both in terms of controlling the behavior of specific word classes and the aggressiveness of the stemmer
	\item it generalizes to unknown words, providing high recall and largely eliminating problems with out-of-vocabulary items
	\item it handles irregular and regular forms with the same mechanism, providing high-precision stemming 
	\item it is simple, utilizing longest matching lookup rather than conditional stemming rules
	\item an open-source reference implementation is provided.
\end{itemize}

\section{XSTEM}
\subsection{Specifying Suffix Rules: An Example}
This section describes how XSTEM works in more detail. The example below considers the different stemming behaviors of words ending in \texttt{-elves}, such as \textit{selves}, \textit{delves}, and \textit{pelves} (the Latin plural form of \textit{pelvis}, which is commonly encountered in biomedical text).

We begin with a very general pattern rule that works for a large number of words, “remove the final \texttt{-s}”. This exemplar is stored in a reverse character trie as shown below.

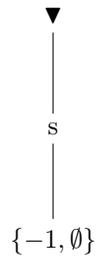
\begin{center}
\begin{tikzpicture}
\node{$\blacktriangledown$} [sibling distance = 2.5cm]
	child {node [] {s}
		child {node {$ \{-1, \emptyset \}$}}
	};
\end{tikzpicture}
\end{center}

A path from the root ($\blacktriangledown$) through the letter \texttt{s} terminates in a suffix operation that specifies truncating the input by one character and appending a null suffix (i.e., drop the final \texttt{-s} and do nothing). This model produces the following output for the three words in our example

\begin{center}
\begin{tabular}{ c c c }
\textit{selves}&$\mapsto$&\textit{selve} \\ 
\textit{delves}&$\mapsto$&\textit{delve} \\ 
\textit{pelves}&$\mapsto$&\textit{pelve} \\  
\end{tabular}
\end{center}

\noindent resulting in two non-words (\textit{selve} and \textit{pelve}) that did not stem to their base forms (\textit{self} and \textit{pelvis}, respectively). 

We add an exemplar that generally works for words ending in \texttt{-lves}, “replace the final \texttt{-ves} with \texttt{-f}”. The rule is added to the trie as shown below.
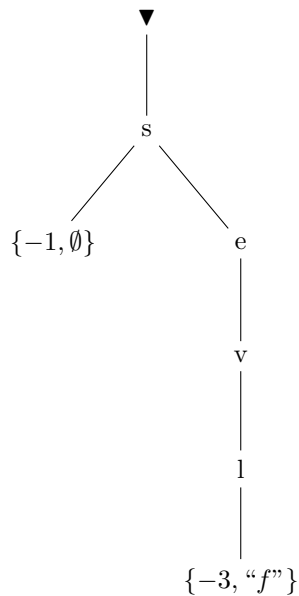
\begin{center}
\begin{tikzpicture}
\node{$\blacktriangledown$} [sibling distance = 2.5cm]
	child {node [] {s}
		child {node {$ \{ -1, \emptyset \}$}}
		child {node {e}
			child { node {v} 
				child { node {l} 
					child {node {$\{ -3, ``f" \}$}}
				}
			}
		}
	};
\end{tikzpicture}
\end{center}

A path from the root through the letters \texttt{s} - \texttt{e} - \texttt{v} - \texttt{l} (reverse traversal of the input) terminates in a suffix operation that specifies truncating the input by three characters and appending a new suffix \texttt{-f}. This model produces the following output for the three words in our example

\begin{center}
\begin{tabular}{ c c c }
\textit{selves}&$\mapsto$&\textit{self} \\ 
\textit{delves}&$\mapsto$&\textit{delf} \\ 
\textit{pelves}&$\mapsto$&\textit{pelf} \\  
\end{tabular}
\end{center}
 
\noindent resulting in two non-words (\textit{delf} and \textit{pelf}) that did not stem to their base forms (\textit{delve} and \textit{pelvis}, respectively). By providing more specific exemplars for \textit{delves} and \textit{pelves}, the model is able to correctly resolve each of the three words in our example to their correct base form, and defaults to \texttt{-s} removal for all other matching words.

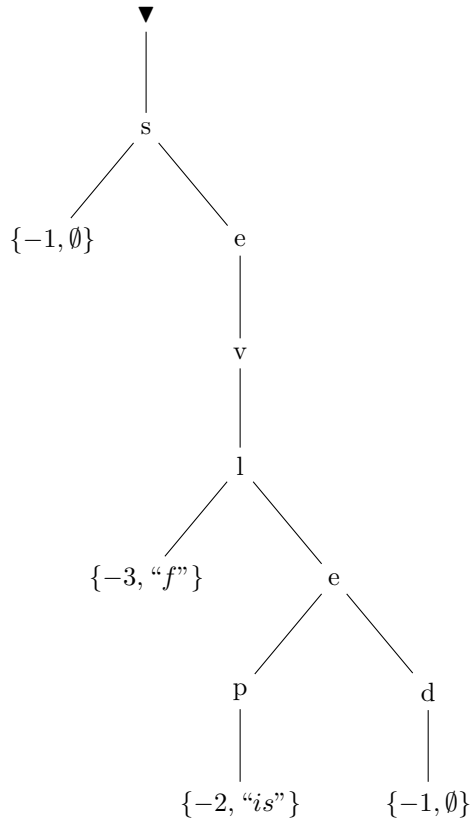
\begin{center}
\begin{tikzpicture}
\node{$\blacktriangledown$} [sibling distance = 2.5cm]
	child {node [] {s}
		child {node {$ \{ -1, \emptyset \}$}}
		child {node {e}
			child { node {v} 
				child { node {l} 
					child {node {$\{ -3, ``f" \}$}}
					child {node {e}
						child {node {p}
							child {node {$\{ -2, ``is" \}$}}
						}
					child {node {d}
							child {node {$\{ -1, \emptyset \}$}}
						}
					}
				}
			}
		}
	};
\end{tikzpicture}
\end{center}

\noindent The updated model produces the following example output

\begin{center}
\begin{tabular}{ c c c }
\textit{selves}&$\mapsto$&\textit{self} \\ 
\textit{delves}&$\mapsto$&\textit{delve} \\ 
\textit{pelves}&$\mapsto$&\textit{pelvis} \\  
\end{tabular}
\end{center}

Although the transformation of word final \texttt{-lves} to singular \texttt{-is} or \texttt{-f} is no longer a productive morphological process in English\footnote{A productive morphological process is one that speakers of a language actively apply to new words.}, such words actively participate in the formation of compound nouns such as \textit{aardwolves}, \textit{beewolves}, \textit{coywolves}, \textit{midpelves}, \textit{hemipelves}, \textit{micropelves}, etc. Other non-productive inflectional patterns, such as the \texttt{-osis/-oses} alternation common in medical terminology derived from Greek, participate even more freely in the formation of novel compound nouns, such as \textit{thrombosis/thromboses}, \textit{microthrombosis/microthromboses}, \textit{dermatosis/dermatoses}, \textit{genodermatosis/genodermatoses}, etc. Examples such as these illustrate the necessity of a stemming model that can generalize beyond a dictionary to handle novel input.

\section{Reference Implementation}
An open-source\footnote{Apache License, Version 2.0} implementation of XSTEM written in Python is available at \url{https://github.com/kirkbaker/xstem}. The implementation does not require any third-party libraries and favors clarity over any particular optimizations for speed or memory, but is nonetheless highly performant. At the time of writing, the reference implementation handles 14 suffix classes, and contains approximately 1500 exemplar suffixes that are distributed across these classes. It also contains an exceptions file (approximately 800 words with irregular morphology)  and a proper names file (approximately 22,000 proper nouns derived
from PubMed author names and geographic location names). Not all of these proper names are strictly required to limit proper suffixation, but are included as a general resource. As with the Krovetz stemmer, it is not strictly necessary to utilize separate exceptions and name files \cite{Kstem}; they are kept distinct for ease of modification.

Each suffix class is stored in a separate trie for configurability; any of these may be omitted from XSTEM in order to produce lighter stemming behavior. Representative stemming modules are described below. 

\paragraph{plural suffixation} In addition to handling English regular and irregular plurals, XSTEM's plural module is notable for the wide range of biomedical Greek and Latin plurals it handles. Examples of Greek vs English plural reduction include alternations such as \textit{osteoscleroses/osteosclerosis} or \textit{aponeuroses/aponeurosis} vs \textit{primroses/primrose} or \textit{rockroses/rockrose}. Examples of Latin plural handling include alternations such as \textit{nanomatrices/nanomatrix} or \textit{oropharynx/oropharynges} vs \textit{prices/price} or \textit{lynx/lynxes}.

\paragraph{\texttt{-er} suffixation}
XSTEM takes a light-handed approach to \texttt{-er} suffixation, focusing on removing the inflectional suffix (e.g., \textit{higher/high} or \textit{healthier/healthy}) but keeping it intact as a derivational suffix (e.g., \textit{renter} $\neq$ \textit{rent} and \textit{reporter} $\neq$ \textit{report}).

\paragraph{\texttt{-ness} suffixation} XSTEM removes \texttt{-ness} when used as a suffix (e.g., \textit{wooziness/woozy} or \textit{boldness/bold}) but not when it can be considered part of the root (e.g., \textit{harness} or \textit{witness}).

\paragraph{\texttt{-ly} suffixation} XSTEM removes adverbial usages of \texttt{-ly} such as \textit{necessarily/necessary} or \textit{steadily/steady}, but not when it is part of the root, such as \textit{apply} or \textit{firefly}.

\paragraph{\texttt{-ity} suffixation} XSTEM also takes a fairly hands-off approach to removing \texttt{-ity}, restricting it to cases where the derived and root forms are closely related, such as \textit{viscosity/viscous} or \textit{obesity/obese}, and keeping it for common roots such as \textit{community} or \textit{quality}.

\paragraph{\texttt{-ize} suffixation} \texttt{-ize} is also removed judiciously in cases of closely related words such as \textit{homogenize/homogenous} or \textit{randomize/random}, and is not removed when doing so would produce a root that is commonly used in a distinct sense. For example, \textit{organize} is not stemmed to \textit{organ} and \textit{polarize} is not stemmed to \textit{polar}. Ultimately, judgments of relatedness are those of the author, and users are encouraged to configure XSTEM to their needs if the defaults are not providing desired results for a given domain.

\paragraph{\texttt{-ing} suffixation} XSTEM is careful to remove only inflectional occurrences of \texttt{-ing}, such as \textit{evading/evade} or \textit{attaining/attain}, while retaining it for roots such as \textit{offspring} or \textit{starling}.

\paragraph{\texttt{-al} suffixation} The suffix \texttt{-al} is removed primarily from closely related words of Latin or Greek origin such as \textit{corneal/cornea} or \textit{esophageal/esophagus} and is kept for words such as \textit{biomaterial} or \textit{fiscal}.

\paragraph{\texttt{-ion} suffixation} As with other derivational suffix modules in XSTEM, \texttt{-ion} removal focuses on alternations that produce closely related words such as \textit{transcription/transcribe} or \textit{inclusion/include}, while preserving distinctions such as \textit{foundation} vs \textit{found} or \textit{portion} vs \textit{port}.

\paragraph{\texttt{-ic} suffixation} XSTEM removes \texttt{-ic} from closely related word pairs such as \textit{algorithmic/algorithm}, \textit{leukemic/leukemia}, \textit{proteomic/proteome}, and \textit{theoretic/theory}, among others. It retains \texttt{-ic} when it is closely associated with the root, such as \textit{mimic}, \textit{epidemic}, or \textit{classic}, for example.

\paragraph{past tense suffixation} The past tense module handles various forms of \texttt{-ed} and \texttt{-d} removal and irregular verbs (and their compounds), such as \textit{misunderstood/misunderstand}, \textit{overtook/overtake}, and the various forms of \textit{be}, among many others. The past tense module does not remove \texttt{-ed} or \texttt{-d} when they are part of the stem, such as \textit{infrared}, \textit{naked}, or \textit{hundred}, etc.

\section{Conclusion}
This paper presented a fast, simple, configurable, high-precision, high-recall stemming algorithm that combines the simplicity and performance of word-based lookup tables with the strong generalizability of rule-based methods to avert problems with out-of-vocabulary words. The model described here overcomes issues that have been previously noted for other approaches, and an open-source reference implementation is available.

\section*{Acknowledgments}
This work was partially supported by the U.S. Department of Defense under a Minerva Initiative Award and the National Institutes of Health (NIH) National Institute on Aging (NIA) under SBIR award 75N95023C00023.

\bibliographystyle{plain}
\bibliography{xstem}

\begin{thebibliography}{10}

\bibitem{Agrawal2022}
Ankit Agrawal, Sarsij Tripathi, Manu Vardhan, Vikas Sihag, Gaurav Choudhary,
  and Nicola Dragoni.
\newblock {BERT}-based transfer-learning approach for nested named-entity
  recognition using joint labeling.
\newblock {\em Applied Sciences}, 12(3), 2022.

\bibitem{BRYCHCIN201568}
Tom\`{a}\v{s} Brychc\'{i}n and Miloslav Konop\`{i}k.
\newblock {HPS}: High precision stemmer.
\newblock {\em Information Processing \& Management}, 51(1):68--91, 2015.

\bibitem{Dawson74}
John Dawson.
\newblock Suffix removal and word conflation.
\newblock {\em ALLC bulletin}, 2(3):33--46, 1974.

\bibitem{Oreilly}
Anand Deshpande and Manish Kumar.
\newblock Dawson stemming.
\newblock In {\em Artificial Intelligence for Big Data}. Packt Publishing,
  2018.
\newblock Accessed 27 April 2022,
  \url{https://www.oreilly.com/library/view/artificial-intelligence-for/9781788472173/c262ef1b-d63a-4f2e-86b9-c3ca08647385.xhtml}.

\bibitem{Hull96}
David~A. Hull.
\newblock Stemming algorithms - a case study for detailed evaluation.
\newblock {\em Journal of the American Society for Information Science},
  47:70--84, 1996.

\bibitem{Jivani2011}
Anjali~Ganesh Jivani.
\newblock A comparative study of stemming algorithms.
\newblock {\em Anjali Ganesh Jivani et al, Int. J. Comp. Tech. Appl}, Vol 2
  (6):1930--1938, 2011.

\bibitem{Johnson2005}
Keith Johnson.
\newblock Decisions and mechanisms in exemplar-based phonology.
\newblock {\em UC Berkeley PhonLab Annual Report}, 1, 2005.

\bibitem{Kstem}
Robert Krovetz.
\newblock Kstem documentation.
\newblock Accessed 27 April 2022,
  \url{http://lexicalresearch.com/kstem-doc.txt}.

\bibitem{Krovetz93}
Robert Krovetz.
\newblock Viewing morphology as an inference process.
\newblock SIGIR '93, page 191–202, New York, NY, USA, 1993. Association for
  Computing Machinery.

\bibitem{Lovins1968}
Julie~B. Lovins.
\newblock Development of a stemming algorithm.
\newblock {\em Mechanical Translation and Computational Linguistics},
  11:22--31, 1968.

\bibitem{Moral2014ASO}
Cristian Moral, Ang{\'e}lica de~Antonio~Jim{\'e}nez, Ricardo Imbert, and Jaime
  Ram{\'i}rez.
\newblock A survey of stemming algorithms in information retrieval.
\newblock {\em Inf. Res.}, 19, 2014.

\bibitem{Paice90}
Chris~D. Paice.
\newblock Another stemmer.
\newblock {\em SIGIR Forum}, 24(3):56–61, Fall 1990.

\bibitem{Porter80}
M.~F. Porter.
\newblock An algorithm for suffix stripping.
\newblock {\em Program: electronic library and information systemsl},
  14:130--137, 1980.

\end{thebibliography}

\end{document}